\documentclass{bmvc2k}
\usepackage[misc]{ifsym}
\usepackage{graphicx}
\usepackage{amsmath}
\usepackage{amssymb}
\usepackage{booktabs}
\usepackage{algpseudocode} 
\usepackage{algcompatible}
\usepackage{algorithm} 
\usepackage{enumerate}
\usepackage{multirow}
\usepackage{makecell}
\usepackage{bmvc2k_natbib}

\title{DiffSketching: Sketch Control Image Synthesis with Diffusion Models}

\addauthor{Qiang Wang}{wanqqiang@bupt.edu.cn}{1}
\addauthor{Di Kong}{dikong@bupt.edu.cn}{1}
\addauthor{Fengyin Lin}{fylin@bupt.edu.cn}{1}
\addauthor{Yonggang Qi$^{\textrm{\Letter}}$}{qiyg@bupt.edu.cn}{1}

\addinstitution{
    Beijing University of Posts and\\
    Telecommunications, Beijing, China
}

\runninghead{WANG, KONG, LIN, QI}{DIFFSKETCHING: SKETCH CONTROL IMAGE SYNTHESIS}

\def\eg{\emph{e.g}\bmvaOneDot}

\def\etal{\emph{et al}\bmvaOneDot}

\begin{document}

\maketitle

\begin{abstract}
Creative sketch is a universal way of visual expression, but translating images from an abstract sketch is very challenging. Traditionally, creating a deep learning model for sketch-to-image synthesis needs to overcome the distorted input sketch without visual details, and requires to collect large-scale sketch-image datasets. We first study this task by using diffusion models. Our model matches sketches through the cross domain constraints, and uses a classifier to guide the image synthesis more accurately. Extensive experiments confirmed that our method can not only be faithful to user's input sketches, but also maintain the diversity and imagination of synthetic image results. Our model can beat GAN-based method in terms of generation quality and human evaluation, and does not rely on massive sketch-image datasets. Additionally, we present applications of our method in image editing and interpolation.
\end{abstract}

\section{Introduction}

\noindent Free-hand sketch is an intuitive way for human beings to express the real world, while imagining from any given sketch to colored realistic images is a desirable ability for intelligent machines. A high quality sketch-to-image synthesis model can help design animation, games and other works. However, sketch contains far less information than image due to its simplicity, abstraction and inaccuracy. The cross-domain synthesis lacks important information such as color, shadow and texture. And the way that people do hand drawing is space distorted and imperfect, which makes this task very difficult.

Early sketch based image synthesis methods \cite{chen2009sketch2photo, eitz2011photosketcher, chen2012poseshop} are based on image retrieval which do not have real generation ability. In recent years, with the rise of GAN \cite{goodfellow2014generative}, a large number of solutions have been proposed \cite{chen2018sketchygan, liu2020unsupervised, huang2018multimodal, wang2021sketch}, but most of these methods rely on large sketch-image pairing datasets, which are very precious and hard to obtain. Sketchy \cite{sangkloy2016sketchy} is the largest sketch-image pairing dataset including 125 categories at present. But each category only contains 50 images, which is far from enough for deep generative models. In addition, almost no research has been attempted on the complex and variable ImageNet \cite{deng2009imagenet} dataset.

Diffusion models rapidly become popular and beat GANs in some key indicators \cite{dhariwal2021diffusion}, bringing new creativity to research the generative models. Inspired by this, we propose DiffSketching, a sketch-guided image synthesis method through diffusion models. The input image is converted into latent noise by forward diffusion process. With the guidance of sketches, adjust the score function to invert to new images. This process does not need to rely on large sketch-image pairing datasets, and can beat the GAN-based method on qualitative comparisons and human evaluation results.

Our goal is that users only need to input a sketch, and our model can generate many corresponding images. There are three main challenges in using diffusion model to complete this task. (i) Existing diffusion models generate data in a single domain, so we need an appropriate guidance method for cross-domain generation, and an appropriate method to measure data distribution in two different domains. (ii) Unlike edge maps extracted from images, sketches and corresponding images are more inconsistent in space and geometry, so it is difficult to measure the cross domain matching of sketch image. (iii) The sketch entered by the user contains little information and often has ambiguity (\eg drawing a dog, it is difficult to tell whether it is a German Shepherd or a Briard). We need to introduce more information to eliminate such ambiguity.

\begin{figure*}
    \centering
    \includegraphics[width=127mm]{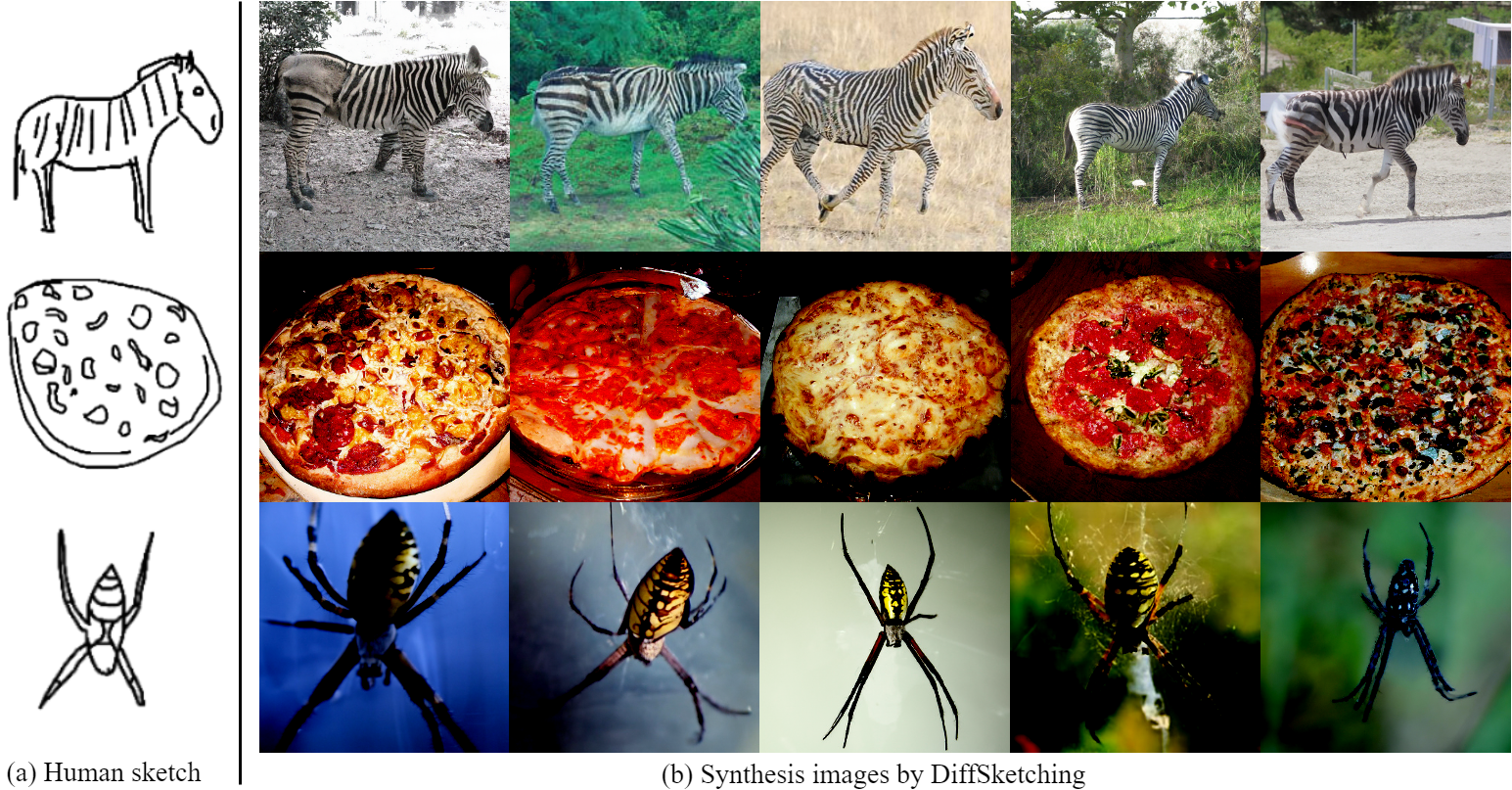}
    \caption{Diffsketching synthesis (b) are a large number of images from (a) one or more real human sketches. Shape, pose, texture and other features of the sketch can be faithfully preserved.}
    \label{fig:all_gneration}
    \vspace{-0.5cm}
\end{figure*}

In order to solve the above challenges, our work makes the following major contributions. (i) We propose a model that can synthesize sketch-faithful, and photo-realistic images from a single sketch (Fig. \ref{fig:all_gneration}), performing better on benchmarks than GAN-based models. (ii) We can guide the generation process more finely and eliminate the singularity and uncertainty of input sketches. (iii) We prove that our method is capable of editing images and conducting image interpolation conditioned on sketch. 

\section{Related works}
{\bf {Sketch Based Image Synthesis}}~~ There are numerous research on edge based image synthesis, which belongs to image translation field \cite{isola2017image, zhu2017unpaired, takano2020generator, xu2020e2i, lin2021edge}. But compared with edges, hand-free sketch is more abstract, imaginative, flexible and challenging. The first work really employ sketch to generate image is SketchyGAN \cite{chen2018sketchygan}, which is an encoder-decoder model and adopts a two stage strategy for shape and appearance completion based on the paired sketch-image data. Subsequently, there has been many works on automatically synthesizing natural images \cite{ghosh2019interactive, gao2020sketchycoco, liu2020unsupervised, wang2021sketch} and human portraits \cite{chen2020deepfacedrawing, li2020deepfacepencil, wu2022deepportraitdrawing}. Most of them are based on GANs, require adversarial training which often suffers from unstableness and mode collapse. 

As a mirror task, image-to-sketch work has also been extensively studied \cite{pang2018deep, song2018learning, li2019photo}. Photo-sketching \cite{li2019photo} trains an image-conditioned contour generator for multiple diverse outputs, achieving the state-of-the-art (SOTA) performance in boundary detection and contour rendering. This method does not generate edge graph, but uses antagonistic training to make the generated result closer to the ground truth hand-drawn sketch. So we use it as a cross domain converter between sketches and images at the stage of measuring perceptual loss.

\noindent
{\bf {Diffusion Models}}~~ Recently, many works on iterative generative models \cite{bengio2014deep}, such as denoising diffusion probabilistic models (DDPM) \cite{ho2020denoising}, score-based generative model \cite{song2020score} can produce samples comparable to those of GANs. Denoising diffusion implicit models (DDIM) \cite{song2020denoising} exert fewer sampling steps to obtain higher quality samples. Prafulla \etal \cite{dhariwal2021diffusion} achieves the SOTA performance in image synthesis by improving DDIM architecture. Because diffusion models do not need adversarial training, they fundamentally solve the mode collapse problem of GANs.

However, a significant drawback of diffusion models is that it simulates many time steps of Markov chain to generate samples. Beyond DDIM, many acceleration methods \cite{lyu2022accelerating, zhang2022fast, watson2021learning, salimans2022progressive, kong2021fast} have been proposed. Besides image synthesis \cite{pandey2022diffusevae, nichol2021glide, rombach2021high, batzolis2021conditional}, diffusion models are widely used in various fields, such as image-to-image translation \cite{li2022vqbb, wolleb2022swiss, choi2021ilvr}, text-to-image translation \cite{jiang2022text2human, saharia2022photorealistic, ramesh2022hierarchical, gu2021vector}, video generation \cite{ho2022video, yang2022diffusion, harvey2022flexible} and audio generation \cite{koizumi2022specgrad, huang2022fastdiff, leng2022binauralgrad}.

\section{Diffusion Models for Sketch-Guidance Image Generation}

\begin{figure*}
    \centering
    \includegraphics[width=\linewidth]{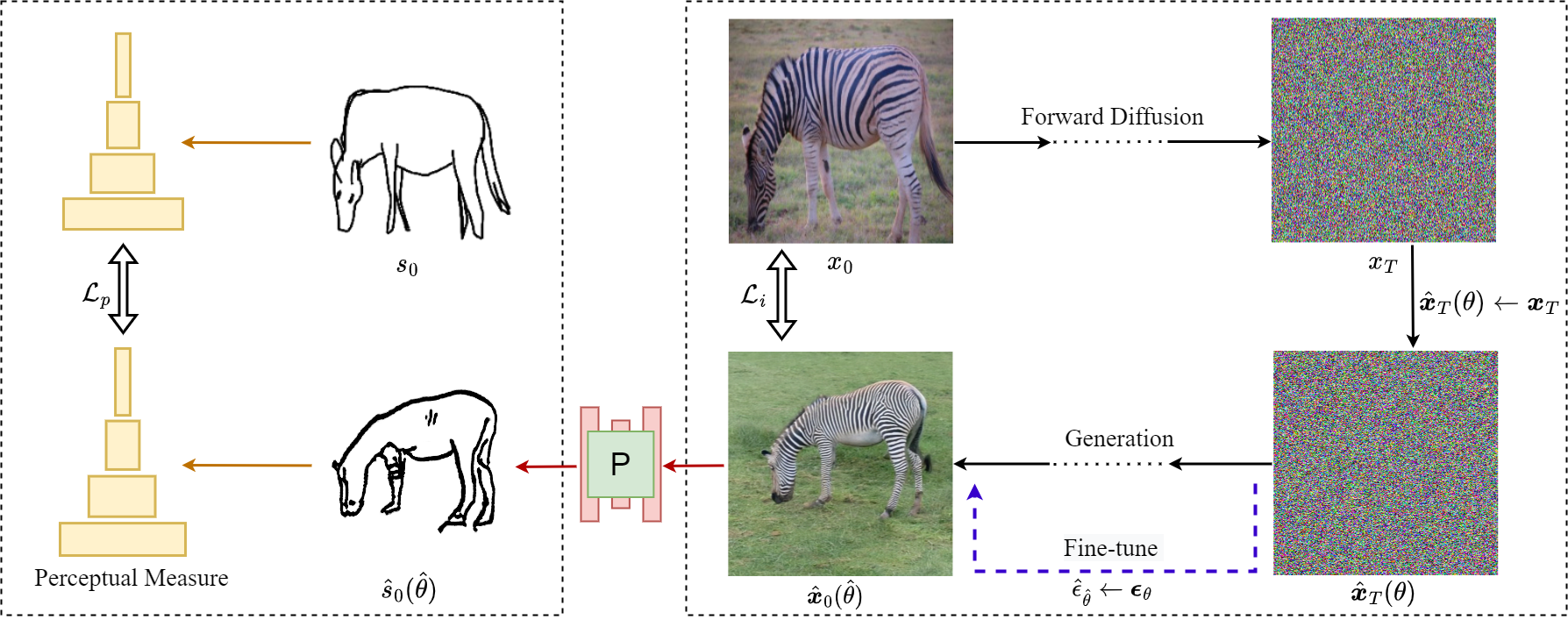}
    \caption{Overview of DiffSketching. Our training Constraints consist two components: (a) $\mathcal{L}_{p}$: the model $\mathcal{P}$ converts $\hat{\boldsymbol{x}}_{0}(\hat{\theta})$ to sketch $\hat s_0(\hat \theta)$ and makes perceptual loss with input sketch $s_0$. (b) $\mathcal{L}_{i}$: cosine image similarity loss between input image $x_0$ and generated image ${\hat x}_{t_{0}}(\theta)$.}
    \label{fig:overview}
    \vspace{-0.6cm}
\end{figure*}

The overview of our proposed DiffSketching's framework is shown in Fig.~\ref{fig:overview}. The input image $x_0$ is converted into latent noise ${x}_{T}$ through the forward diffusion process. We clone $x_T$ to $\hat x_T(\theta)$ and then synthesize the image ${\hat x}_{0}(\theta)$ from $\hat x_T(\theta)$ via a reverse generation process which is achieved through a fine-tuning process.

\subsection{Background}

\noindent 
{\bf {Forward Diffusion Process}}~~ Diffusion models slowly inject noise into the original data to destroy the initial data distribution. During the reverse generation process, the probability distribution of the desired data $\hat x_0$ is obtained by learning to predict the noise and denoising. For the distribution of each training data ${x_0}\sim q_{data}(x_0)$, through a variance schedule $\beta_{1}, \dots, \beta_{T}$, diffusion models gradually add Gaussian noise $\epsilon$ in step $t$ to get $x_1, \dots ,x_{T}$:

\vspace{-0.3cm}
\begin{equation}
\label{eq:fwd}
    q(\textbf{x}_{1:T}|\textbf{x}_0) := \prod_{t=1}^T q(\textbf{x}_t|\textbf{x}_{t-1})
\end{equation}
\vspace{-0.3cm}

Ho \etal~\cite{ho2020denoising} used $\alpha_{t}:=1-\beta_{t}$ and $\bar{\alpha}_{t}=\prod_{i=1}^{T} \alpha_{i}$ to represent  $\mathbf{x}_{t}\left(\mathbf{x}_{0}, \boldsymbol{\epsilon}\right)=\sqrt{\bar{\alpha}_{t}} \mathbf{x}_{0}+\sqrt{1-\bar{\alpha}_{t}} \boldsymbol{\epsilon}$, where $\boldsymbol{\epsilon} \sim \mathcal{N}(0, \boldsymbol{I})$. Song \etal~\cite{song2020denoising} proposed DDIM that changed forward Markov process to Non-Markov process by using variable information. It becomes an implicit probabilistic model:

\vspace{-0.3cm}
\begin{equation}
\label{eq:q_sigma}
    q_{\sigma}(\boldsymbol x_{t}|\boldsymbol x_{t-1},\boldsymbol x_{0})=\frac{q_{\sigma}(\boldsymbol x_{t-1}|\boldsymbol x_{t},\boldsymbol x_{0})q_{\sigma}(\boldsymbol x_{t}|\boldsymbol x_{0})}{q_{\sigma}(\boldsymbol x_{t-1}|\boldsymbol x_{0})}
\end{equation}
\vspace{-0.3cm}

where $\sigma\in\mathbb{R}_{\mathrm{{\geq 0}}}^{T}$ is the index of inference distribution family $\mathcal Q$, controlling the stochasticity of the forward process. 

\noindent 
{\bf {Reverse Generation Process}}~~ In the reverse generation process $p_\theta(x_t)$, diffusion models allow for different reverse samples to be generated by varying the variance of noise. It establishes the mapping relationship from latent to image and conducts denoising from $x_t$ to get $x_{t-1}$: 

\vspace{-0.3cm}
\begin{equation}
\begin{aligned}
\label{eq:x_t-1_2}
    \boldsymbol{x}_{t-1}=\sqrt{\frac{\alpha_{t-1}}{\alpha_{t}}} \boldsymbol{x}_{t}+\left(\sqrt{1-\alpha_{t-1}}-\sqrt{\frac{{\alpha_{t-1}}(1-\alpha_t)}{\alpha_{t}}}\right) \boldsymbol{\epsilon}_{\theta}\left(\boldsymbol{x}_{t}, t\right)
\end{aligned}
\end{equation}
\vspace{-0.3cm}

The function $\epsilon_{\theta}(x_t,t)$ represents the prediction of noise distribution, and $\theta$ denotes the learnable parameter. Training process randomly samples the image with noise in time step $t$, and adopts simple mean squared error loss to make predicted noise closer to true noise: $\nabla_{\theta}\left\|\epsilon-\epsilon_{\theta}(\sqrt{\alpha_{t}}\mathbf{x}_{0}+\sqrt{1-{\alpha_{t}}}\epsilon,t)\right\|^{2}$.

\noindent
{\bf {Classifier Guidance}}~~ Prafulla \etal~\cite{dhariwal2021diffusion} adopted classifier to guide the generation of images which does not need additional training. This method directly generates the desired image through the gradient guidance of the trained external classifier $p_{\phi}(y|x_{t}, t)$ on the trained diffusion models, where $y$ is the class label. The predicted noise is defined as:

\vspace{-0.3cm}
\begin{equation}
\label{eq:theta_t}
    \hat{\epsilon}_{\theta}(x_{t}, t) = \epsilon_{\theta}(x_{t}, t)-\sqrt{1-\bar{\alpha}_{t}}\,\nabla_{x_{t}}\log p_{\phi}(y|x_{t})
\end{equation}
\vspace{-0.3cm}

During sampling, the sampling center of the expected noisy image $x_t$ is guided by the classifier to the condition that the predicted noise is as close as possible to the true noise and can guide the reverse diffusion direction to the desired category. 

\subsection{Perceptual Diversity Learning} \label{Perceptual Diversity Learning}

\noindent We define $\mathcal{X}$, $\mathcal{Y}$ as the domains of sketches and images respectively. To bridge the gap between  $\mathcal{X}$ and $\mathcal{Y}$, we adopt the pretrained GAN-based network Photo-sketching \cite{li2019photo} to translate images into sketches $\mathcal{P}:{\hat x_0}({\hat \theta})\rightarrow {\hat s_0}(\hat \theta)$. 

Because sketches have strong space distortion and style variability, classical per-pixel measurement methods such as $\ell_{1}$ Manhattan Distance or $\ell_{2}$ Euclidean Distance \cite{dokmanic2015euclidean} will greatly damage the diversity of the generated sketches and enlarge the input defect which misguides the model. Therefore, we introduce perceptual metric loss \cite{zhang2018unreasonable} which can express appearance similarity from global semantics to solve this problem. We use a pretrained perceptual sketch feature extractor $F_s(\cdot)$ for feature extraction between $s_{0}$ and ${\hat s_0}(\hat \theta)$. $w_l$ is denoted to scale the activation channel-wise for each layer $l$. Then we calculate $l_2$ distance, average over space and sum over channel wise:

\vspace{-0.3cm}
\begin{equation}
\label{eq:l_p}
    \mathcal L_{p}=\sum_{l=1}^{L}\frac{1}{H_{l}W_{l}}\sum_{h,w}||w_{l}\odot(F_s(\hat s_0(\hat \theta))_{h w}^{l}-F_s(s_0)_{hw}^{l})||_{2}^{2}
\end{equation}

where $w_{l} \in \mathbb{R}^{C_l}$ and $F_s(\hat s_0), F_s(s_0)\in \mathbb{R}^{H_{l}\times W_{l}\times C_{l}}$.

\subsection{Image Constraint Identity Learning} \label{Image Constraint Identity Learning}

\begin{figure*}
    \centering
    \includegraphics[width=120mm]{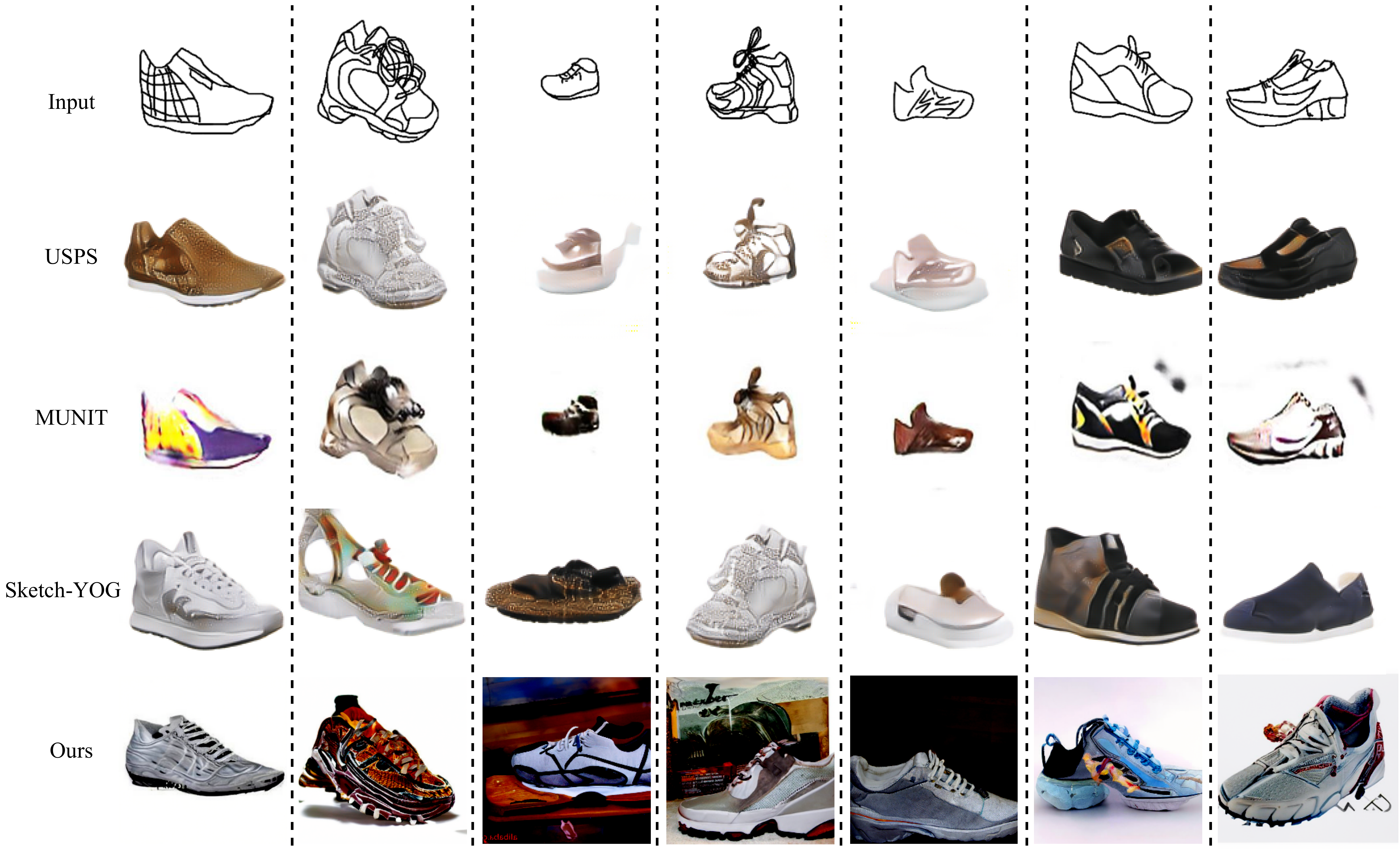}
    \caption{Qualitative results compared with baselines under the same sketch input.}
    \label{fig:shoe_compare}
    \vspace{-0.5cm}
\end{figure*}

\begin{figure*}
    \centering
    \includegraphics[width=120mm]{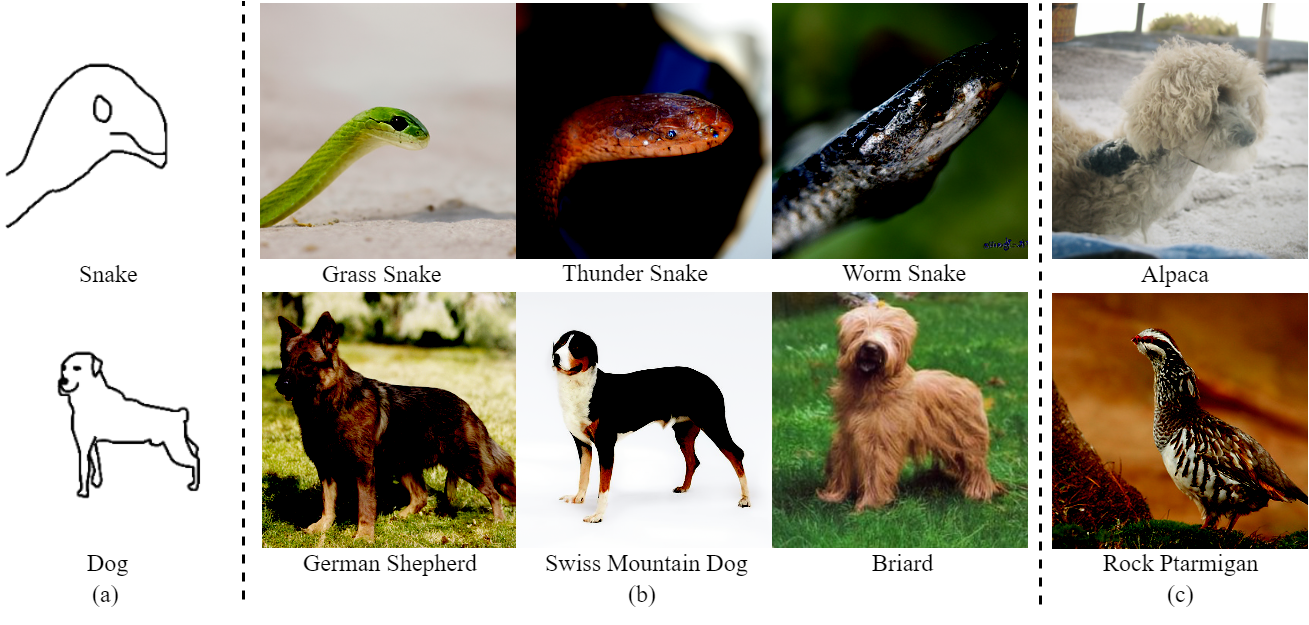}
    \caption{Fine-grained sketch controlling. (a) is input sketch, (b) is a fine-grained category that can be specified by users, (c) is a category that independent of input sketch.}
    \label{fig:fine-grained}
    \vspace{-0.5cm}
\end{figure*}

\noindent
We observe that only the constraints in the sketch domain will lead to a loss of too many elements in the original image and an increase in generation uncertainty. Because the sketch domain provides less information than the image domain. To solve this problem, we propose an image constraint identity loss to compare the input image with the generated one.

We trained ResNet-50 \cite{he2016deep} as the image constraint feature extractor $F_i$ to extract features from $x_0$ and ${\hat x}_{0}(\hat \theta)$ in an attempt to minimize the cosine distance of the generated image from the input image:

\vspace{-0.3cm}
\begin{equation}
\label{eq:l_i}
    \mathcal L_i=\frac{\mathbf{F_i(x_0) \cdot F_i({\hat x}_{0}(\hat \theta))}}{\|\mathbf{F_i(x_0)}\|\|\mathbf{F_i({\hat x}_{0}(\hat \theta))}\|}
\end{equation}

Image constraint identity loss enables the generated results to have more identity information and enhances the robustness of the model. Sketch perceptual loss adds diversity and imagination. $\lambda$ is a super parameter for balancing diversity and identity. Our training objective is as follows:

\vspace{-0.3cm}
\begin{equation}
\label{eq:l}
    \mathcal{L}=\lambda\mathcal L_{i}(x_0, {\hat x}_{0}(\hat \theta)))+(1-\lambda)\mathcal{L}_{p}(s_0, \hat s_0(\hat \theta))
\end{equation}

\subsection{Class-Guidance Fine-tuning Reverse Process} \label{Class-Guidance Fine-tuning Reverse Process}
According to Eq.~\ref{eq:x_t-1_2}, the backward generation process is denoising from $\hat x_ T(\theta)$ to $\hat x_0(\theta)$. The sketch drawn with a few strokes is too simple and can easily mislead the model to generate inaccurate results. To prevent the generated data distribution from deviating from the category center, we constrain the model by a classifier, via the Eq.~\ref{eq:theta_t}. 

To take full advantage of the image synthesis performance of diffusion models, we pre-train the forward and reverse process of diffusion models with a classifier. In terms of fine-tuning, our model learns to be self-supervised subject to the constraint of Eq.~\ref{eq:l}. Once the diffusion model has been fine-tuned, any input sketches can be processed into images, as shown in Fig.~\ref{fig:all_gneration}. More details on the fine-tuning procedure and the structure of the model are analyzed in the supplementary materials.

\section{Experiments}
\subsection{Evaluations}

\noindent
{\bf {Datasets}}~~ We select ImageNet dataset with $256 \times 256$ resolution, including 128K images of 1000 categories, to pretrain class-guidance diffusion models. In the fine-tuning stage, we take 12.5K images from Sketchy \cite{sangkloy2016sketchy} dataset with each image corresponding to $5\sim 10$ pieces of sketches.

\noindent
{\bf {Quantitative evaluation}}~~ We measure our model sample quality based on Fréchet Inception Distance \cite{heusel2017gans} (FID), Inception Score \cite{barratt2018note} (IS) , Precision and Recall metrics \cite{kynkaanniemi2019improved}. FID measures the distribution similarity between real images and generated images by comparing the mean and variance of image features. IS calculates the classification entropy of the generated image distribution. The Precision measures fidelity that the model samples are close to the data samples in VGG feature space \cite{simonyan2014very}, and the Recall measures diversity that the data samples are close to the model samples in VGG feature space.

\noindent
{\bf {Human study}}~~ We conduct human study to judge the synthetic quality by comparing the output of different baseline methods with our method. Given an input sketch and the output of different methods, participants are asked to select the image that best conforms to the characteristics of sketch in the output. A total of 10 viewers were recruited for this test. We randomly selected 500 samples which were randomly displayed. And the percentage of each selected method was counted.

\begin{figure*}
    \centering
    \includegraphics[width=120mm]{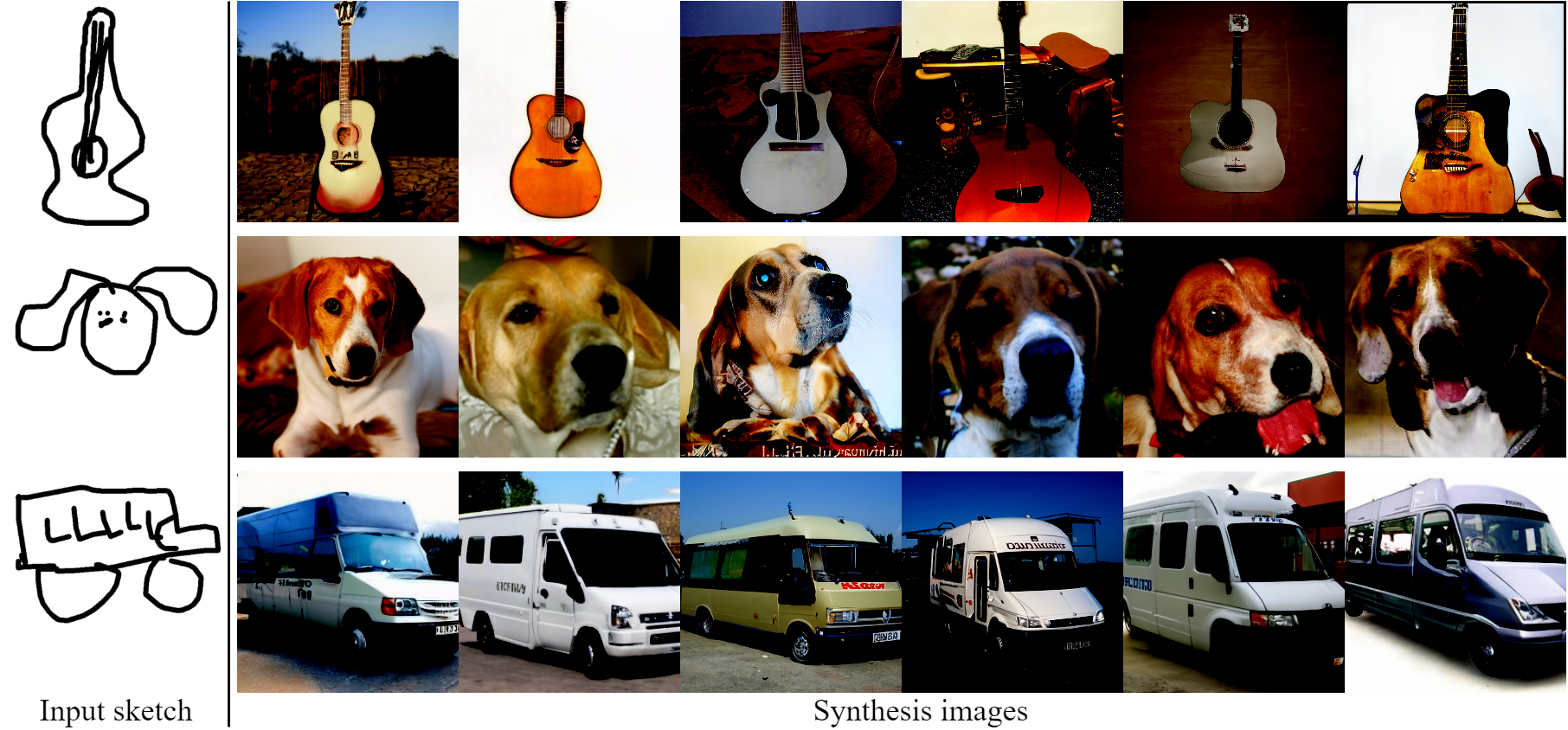}
    \vspace{-0.2cm}
    \caption{ Qualitative results testing on Quickdraw.}
    \label{fig:test_quickdraw}
\end{figure*}


\subsection{Comparison}
\noindent {\bf Baselines}~~ To the best of our knowledge, this is the first time diffusion models have been used for sketch-based image synthesis and most of the previous works are based on GANs. We choose 3 baseline models and to be fair, all methods are tested on Sketchy evaluation dataset. (i) {USPS \cite{liu2020unsupervised}} is an unsupervised GAN model consisting of two steps, translating the sketch shape into a gray-scale image and enriching it into a color image. It proposes an attention module to deal with abstraction and style variations which can improve the quality and realism of generation. (ii) {MUNIT \cite{huang2018multimodal}} is a general unsupervised multimodal image translation framework. MUNIT decomposes the image into a content code and a style code. It recombines the content code with the randomly sampled style code. And the model learns both codes at the same time. (iii) {Sketch-YOG \cite{wang2021sketch}} pretrains a GAN-based generation model, utilizing cross domain adversarial learning and image space regulation to fine-tune.

\begin{table} 
  \caption{Quantitative result. The best value is highlighted in black.}
  \centering
  \begin{tabular}{ccccccc}
    \toprule
    Method & \thead{FID \\ $\downarrow$} & \thead{IS \\ $\uparrow$} & \thead{Precision \\ $\uparrow$} & \thead{Recall \\ $\uparrow$} & \thead{Human \\ $\uparrow$} \\
    \midrule
    USPS & 48.73 & 23.74 & 0.42 & 0.38 & 26.45\% \\
    MUNIT & 56.50 & 28.99 & 0.34 & 0.51 & 20.23\% \\
    Sketch-YOG & 19.94 & 48.94 & \textbf{0.70} & 0.53 & 18.85\% \\
    \midrule
    Ours & \textbf{6.46} & \textbf{89.91} & 0.68 & \textbf{0.56} & \textbf{34.47\%} \\
    \midrule
    Ours (w/o $\mathcal{L}_p$) & 7.22 & 83.43 & 0.33 & 0.39 & N/A \\
    Ours (w/o $\mathcal{L}_i$) & 11.78 & 63.09 & 0.40 & 0.44 & N/A \\
    Ours (Quickdraw) & 6.65 & 87.42 & 0.67 & 0.49 & N/A \\ 
    \bottomrule
  \end{tabular}
  \label{quantitative comparison}
\end{table}

\noindent
{\bf {Benchmarking and Qualitative results}}~~ (i) Our model outperforms other baselines in almost all metrics listed in Table \ref{quantitative comparison}, indicating that we can restore images with more diversity and high fidelity. The higher human study score shows that our synthetic results are more in line with human intuitive judgments. (ii) Unlike USPS and MUNIT, our model does not need large-scale sketch image datasets. Due to the lack of such large datasets, many GAN-based sketch-to-image models can only synthesize a few categories such as shoes and chairs. We compared qualitative results on shoes, shown in Fig. \ref{fig:shoe_compare}. (iii) USPS and MUNIT generate image shapes that strictly match the input sketches and they focus on the generation of color, texture and shading. Whereas Sketch-YOG and our method give the model more imagination in terms of external contours, in particular we are able to generate more complex backgrounds, resulting in a higher IS score. (iv) The Precision is slightly lower than that of Sketch-YOG, indicating that our model is slightly less sensitive to the distribution of real data. More comparison results and analysis can be found in supplementary material.

\subsection{Sketch-Based Image Synthesis}

\noindent
{\bf {Fine-grained sketching controls image synthesis}}~~ As shown in Fig.~\ref{fig:fine-grained}(a), when sketching a dog, the user' s simple strokes could not be identified as German Shepherd, Briard, Swiss Mountain Dog or any other categories. The trained classifiers enable users to specify the categories they want to generate in Fig.~\ref{fig:fine-grained}(b). More interestingly, when we specify categories that are not related to the original input sketch, we can still synthesize images that are similar in style and shape to the sketch, as displayed in Fig.~\ref{fig:fine-grained}(c).

\noindent
{\bf {Test on real human sketches}}~~ To prove our model is more practical than other methods, we conduct tests on hand-drawn sketches. Quickdraw \cite{ha2017neural} collects real hand-painted sketches, which is simple and distorted. Qualitative and quantitative results are shown in Fig.~\ref{fig:test_quickdraw} and Table~\ref{quantitative comparison}, respectively. Since we calculate the global perception loss of sketch rather than the one-to-one pairing of local details, our model can still generate the user's ideal results from realistic and distorted input sketches with only slightly reduced quantitative indicators. 

\begin{figure*}
    \centering
    \includegraphics[width=\linewidth]{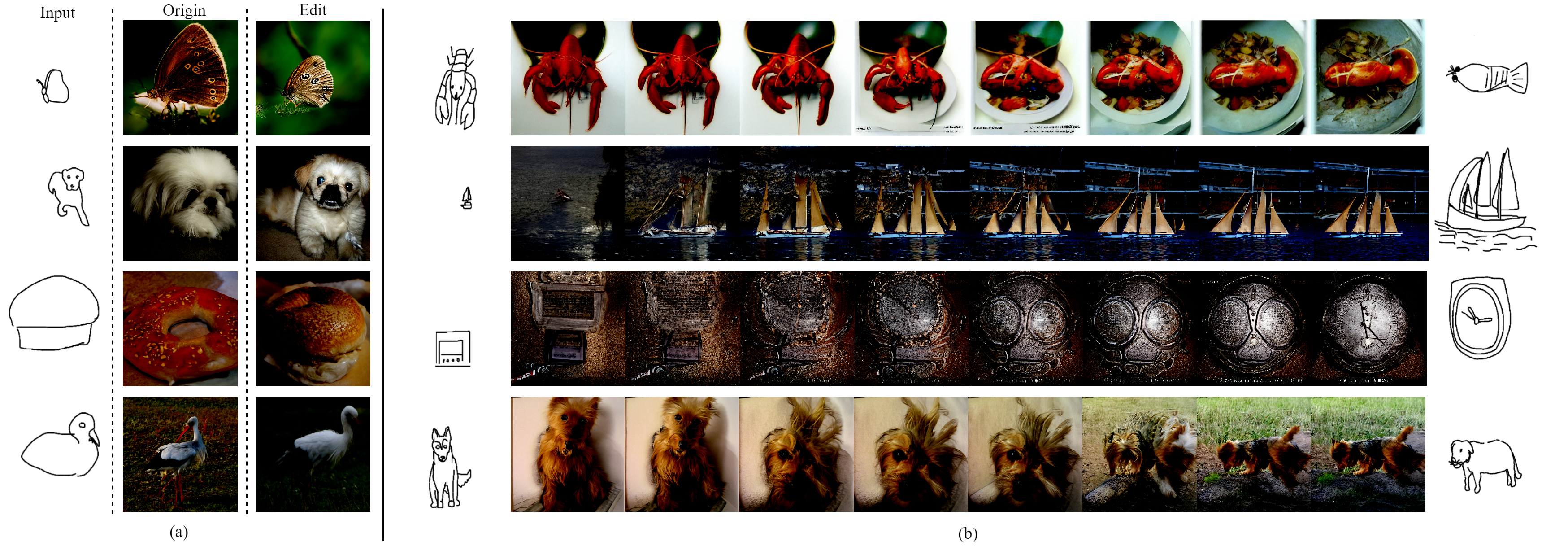}
    \vspace{-0.5cm}
    \caption{(a) Image editing. (b) Condition Interpolation.}
    \label{fig:image edit and interpolation}
    \vspace{-0.5cm}
\end{figure*}

\subsection{Applications}

\noindent
{\bf {Image editing}}~~ Our model can edit the original images under the guidance of input sketches to obtain new images. We use the Attentional block in Luhman \etal \cite{luhman2020diffusion} to input the origin image information, retaining the basic texture and color information of the picture. And we modify the posture and shape features according to the user's input sketch to synthesize a new image. The results are shown in Fig.~\ref{fig:image edit and interpolation}(a). Details about this method are provided in supplementary material.

\noindent
{\bf {Condition Interpolation}}~~ Because of the consistency of DDIM, we use spherical linear \cite{shoemake1985animating} to combine different initial latent variables $x_T^{(0)}$ and $x_{T}^{(1)}$ to get a new $x_T^{(\alpha)}$.

\vspace{-0.3cm}
\begin{equation}
\begin{aligned}
\label{eq:interpolation}
    x_{T}^{(\alpha)}=\frac{\sin((1-\alpha)\theta)}{\sin(\theta)}x_{T}^{(0)}+\frac{\sin(\alpha\theta)}{\sin(\theta)}x_{T}^{(1)}
\end{aligned}
\end{equation}

where $\theta=\mathrm{arccos}\left({\frac{(x_{T}^{(0)})^{\top}x_{T}^{(1)}}{||x_{T}^{(0)}|||x_{T}^{(1)}||}}\right)$, $\alpha \sim (0, \pi/2)$. We linearly extract eight $\alpha$ values, and show the results in Fig. \ref{fig:image edit and interpolation}(b). The left and right sketches are different inputs. Between them are reconstructed interpolation results in latent space. More results of interpolation method are shown in supplementary material.

\subsection{Ablations}
\begin{figure*}
    \centering
    \includegraphics[width=120mm]{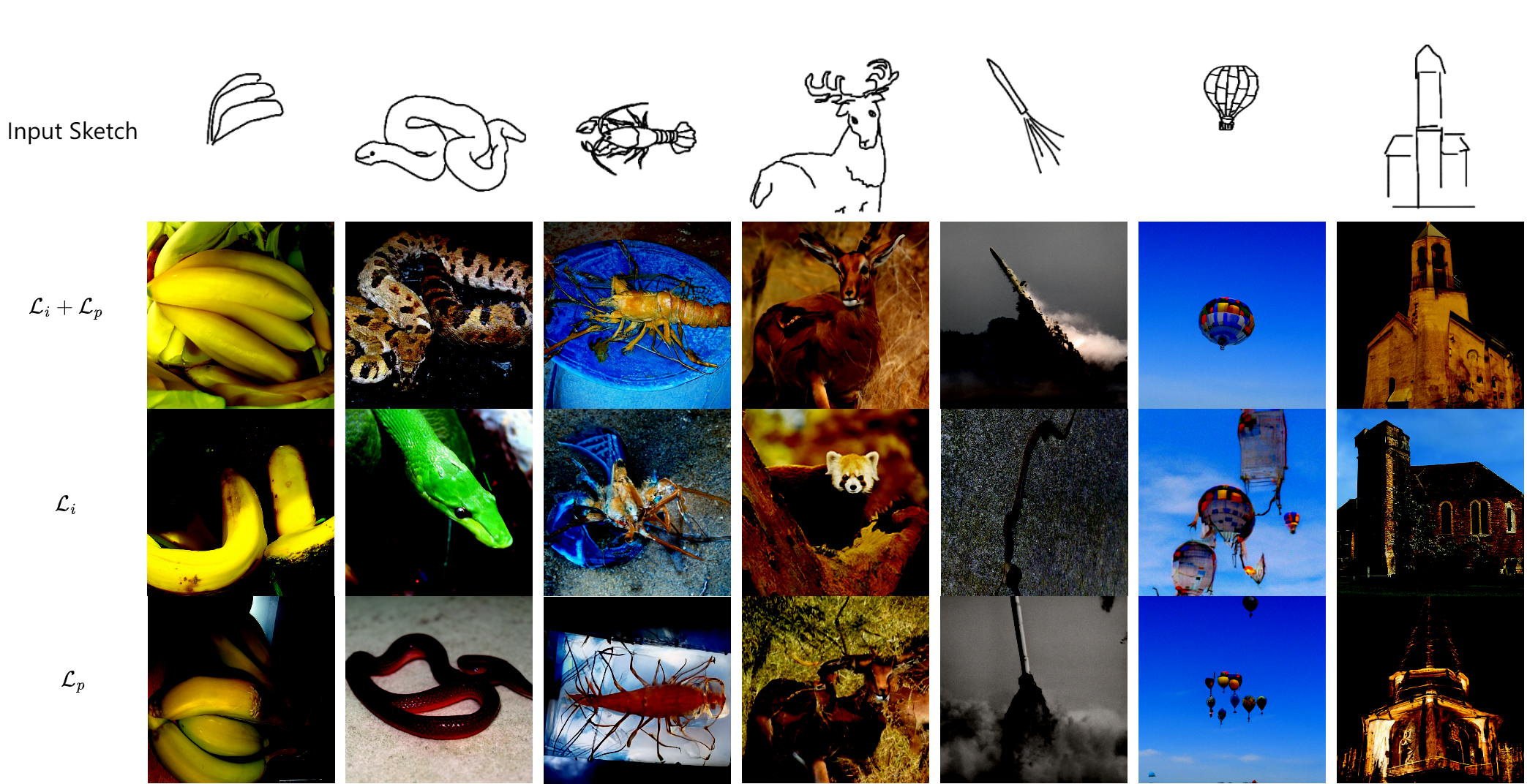}
    \caption{Ablation study of identity loss $\mathcal L_{i}$ and perceptual loss $\mathcal L_{p}$.}
    \label{fig:ablation_figure}
    \vspace{-0.5cm}
\end{figure*}

\noindent From Fig.~\ref{fig:ablation_figure}, both image identity loss $\mathcal L_{i}$ and sketch perceptual loss $\mathcal L_{p}$ are the keys to the success of our model. (i) As shown in Table~\ref{quantitative comparison}, the absence of either $\mathcal L_{p}$ or $\mathcal L_{i}$ degrades the quality of the synthesis. (ii) Without introducing $\mathcal L_{p}$, the model is unable to guide the generation process of image. And the synthesis result is not associated with input sketch. (iii) Without introducing $\mathcal L_ {i} $, although the model can still recreate the general shape and position of the sketch, a great deal of identity and detail information will be lost, making the generated image feel vague in texture. 

\section{Conclusion}

We propose DiffSketching, the first cross-domain sketch-to-image synthesis method utilizing diffusion models. Our method can be self-supervised when matching inputs, overcoming the large domain gap between sketch and generator's parameter space. We can distinguish sketches of simple lines through the classifier, showing strong content inference ability. And the DiffSketching outperforms GAN-based methods on many key metrics, achieving high-quality and realistic results. We further show the potential for application to other tasks, such as image editing and condition interpolation. 

\bibliography{egbib}
\end{document}